\documentclass[10pt,twocolumn,letterpaper]{article}
\usepackage{iccv}     

\usepackage[dvipsnames]{xcolor}
\usepackage{amsfonts}
\usepackage{amsmath}
\usepackage{amssymb}
\usepackage{booktabs}
\usepackage{graphicx}
\usepackage{microtype}
\usepackage{nicefrac}
\usepackage{url}
\usepackage{xcolor}
\usepackage{xspace}
\usepackage{multirow}

\definecolor{cvprblue}{rgb}{0.21,0.49,0.74}
\usepackage[pagebackref,breaklinks,colorlinks,allcolors=cvprblue]{hyperref}

\def\paperID{12633}
\def\confName{ICCV}
\def\confYear{2025}

\newcommand{\method}{DPM\xspace}
\newcommand{\longmethod}{Dynamic Point Maps\xspace}
\newcommand{\duster}{DUSt3R\xspace}
\newcommand{\monster}{MonST3R\xspace}
\newcommand{\greencheck}{\textcolor{green}{\checkmark}}
\newcommand{\bluecheck}{\textcolor{blue}{\checkmark}}
\newcommand{\magcheck}{\textcolor{magenta}{\checkmark}}

\makeatletter
\renewcommand{\paragraph}{%
  \@startsection{paragraph}{4}%
  {\z@}{-0.5em}{-0.5em}%
  {\normalfont\normalsize\bfseries}%
}
\newcommand{\tablestyle}[2]{\setlength{\tabcolsep}{#1}\renewcommand{\arraystretch}{#2}\centering\footnotesize}

\makeatother

\title{\longmethod: A Versatile Representation for Dynamic 3D Reconstruction}

\author{Edgar Sucar \hspace{1em} Zihang Lai \hspace{1em} Eldar Insafutdinov \hspace{1em} Andrea Vedaldi\\
Visual Geometry Group (VGG), University of Oxford\\
{\tt\small \{edgarsucar,zlai,eldar,vedaldi\}@robots.ox.ac.uk}
}

\begin{document}

\twocolumn[{
\maketitle
\vspace{2em}
\begin{center}
\includegraphics[width=\linewidth]{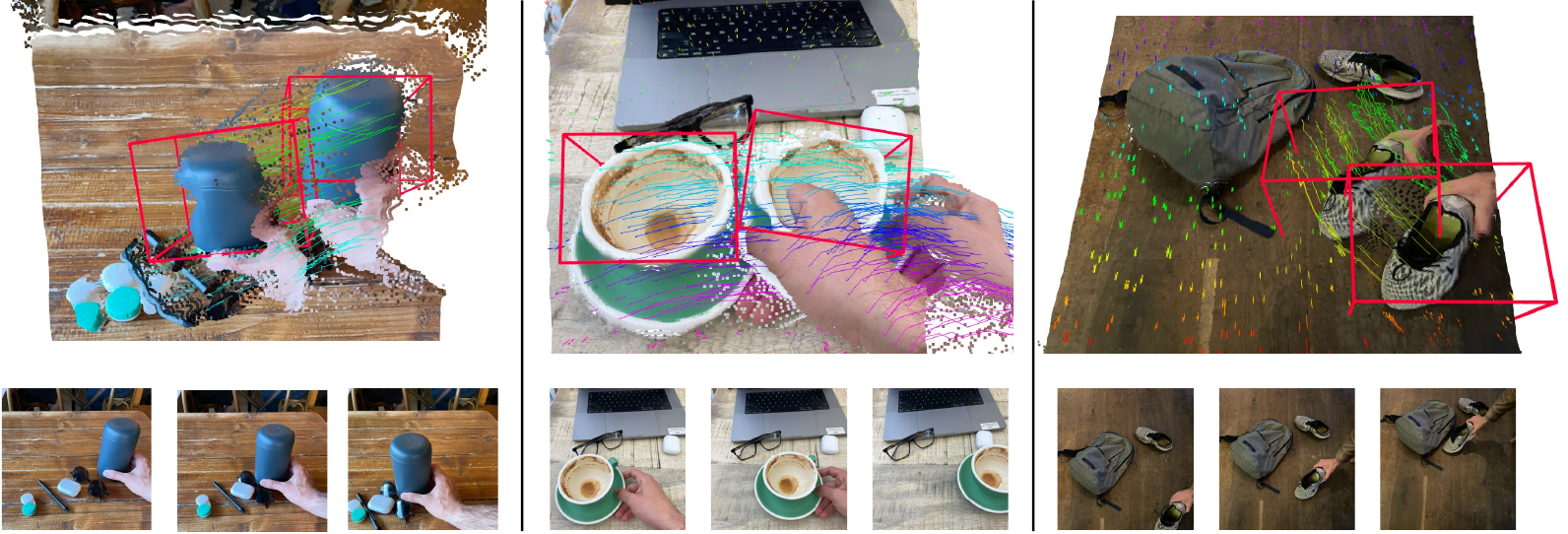}
\end{center}
\vspace{-1em}
\captionsetup{type=figure}
\captionof{figure}{We introduce the concept of \textbf{Dynamic Point Maps} (\method).
Differently from previous extension of point maps to dynamics, \method{s} are \emph{time invariant} in addition to be viewpoint invariant.
Because of this, by predicting \method{}s from a pair of images in a feed-forward manner we can easily solve key 3D tasks, such as recovering the camera parameters and reconstructing shape, as well 4D ones, such as estimating scene flow and 3D object tracking.
}%
\label{fig:teaser}
\vspace{2em}
}]

\begin{abstract}

\duster has recently shown that one can reduce many tasks in multi-view geometry, including estimating camera intrinsics and extrinsics, reconstructing the scene in 3D, and establishing image correspondences, to the prediction of a pair of viewpoint-invariant point maps, i.e., pixel-aligned point clouds defined in a common reference frame.
This formulation is elegant and powerful, but unable to tackle dynamic scenes.
To address this challenge, we introduce the concept of \longmethod (\method), extending standard point maps to support 4D tasks such as motion segmentation, scene flow estimation, 3D object tracking, and 2D correspondence.
Our key intuition is that, when time is introduced, there are several possible spatial and time references that can be used to define the point maps.
We identify a minimal subset of such combinations that can be regressed by a network to solve the sub tasks mentioned above.
We train a \method predictor on a mixture of synthetic and real data and evaluate it across diverse benchmarks for video depth prediction, dynamic point cloud reconstruction, 3D scene flow and object pose tracking, achieving state-of-the-art performance. Code, models and additional results are available at \href{https://www.robots.ox.ac.uk/~vgg/research/dynamic-point-maps/}{https://www.robots.ox.ac.uk/\textasciitilde vgg/research/dynamic-point-maps/}.
\end{abstract}    
\section{Introduction}%
\label{sec:intro}

\begin{figure*}
\input{sec/fig_model}
\end{figure*}
\begin{figure*}[htb]
\centering
\footnotesize
\begin{minipage}[c]{9.8em}
$P_1({\color{magenta}t_1},\pi_1)$%
\begin{minipage}[c]{5em}
\includegraphics[width=\textwidth,clip,trim={0 10em 0 0}]{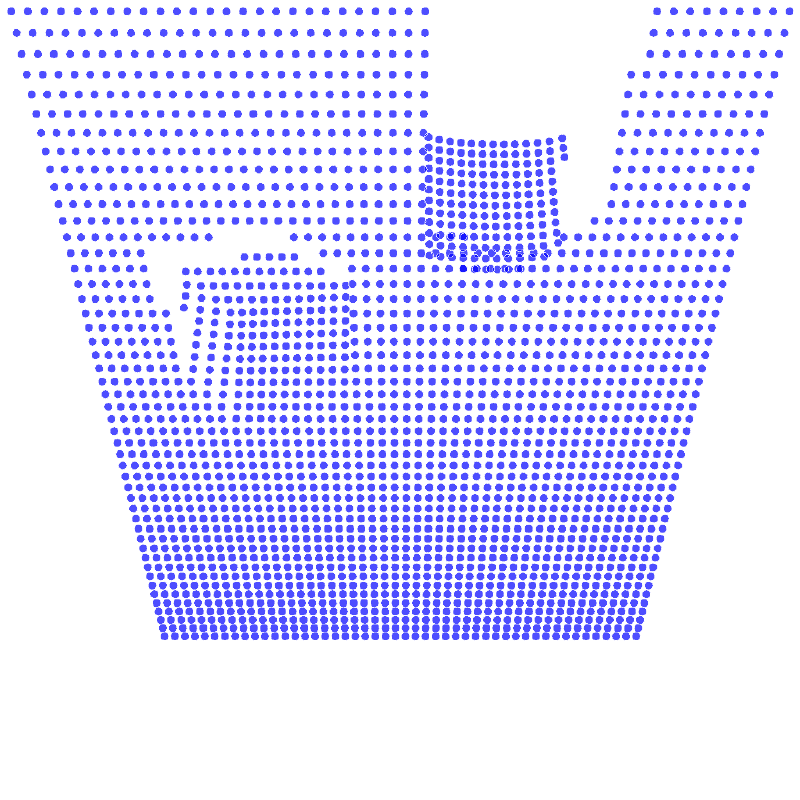}%
\end{minipage}
\\
$P_2({\color{teal}t_2},\pi_1)$%
\begin{minipage}[c]{5em}
\includegraphics[width=\textwidth,clip,trim={0 10em 0 0}]{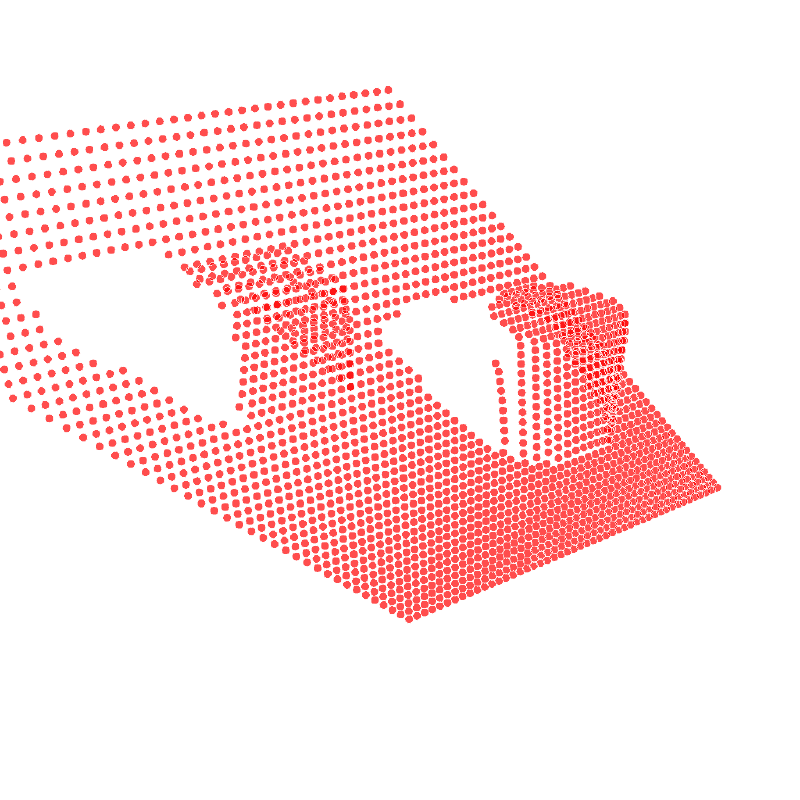}%
\end{minipage}
\end{minipage}%
~$\Rightarrow$~%
\begin{minipage}[c]{10em}\centering
\includegraphics[width=\textwidth,clip,trim={0 10em 0 0}]{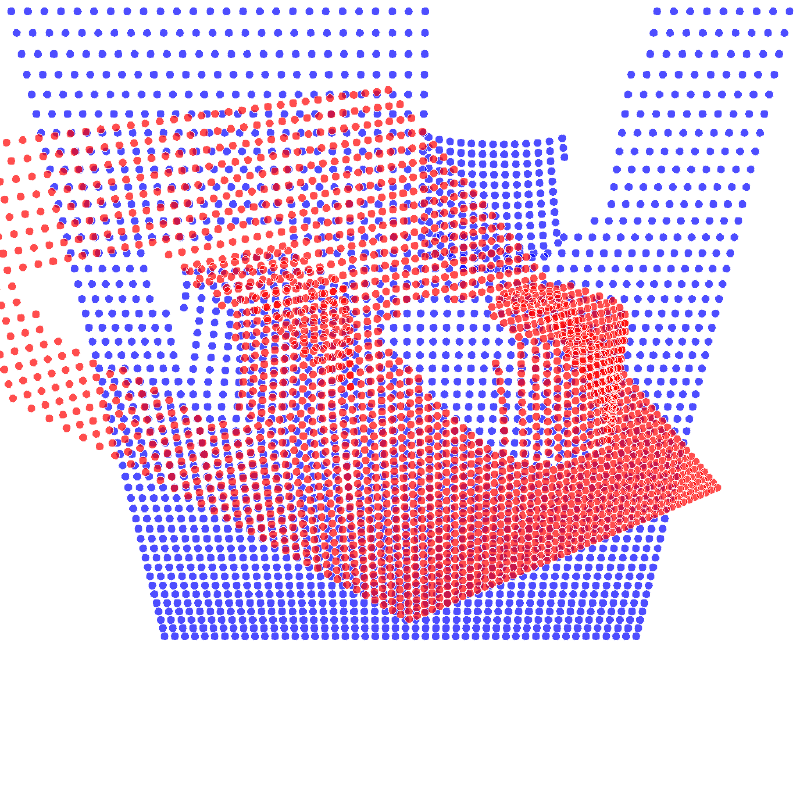}
${\color{magenta}t_1}\not={\color{teal}t_2}$%
\end{minipage}%
~~\vrule~\vrule~~%
\begin{minipage}[c]{19em}
$P_1({\color{magenta}t_1},\pi_1)$%
\begin{minipage}[c]{5em}
\includegraphics[width=\textwidth,clip,trim={0 10em 0 0}]{Figures/P1_t1.pdf}
\end{minipage}%
$P_1({\color{teal}t_2},\pi_1)$%
\begin{minipage}[c]{5em}
\includegraphics[width=\textwidth,clip,trim={0 10em 0 0}]{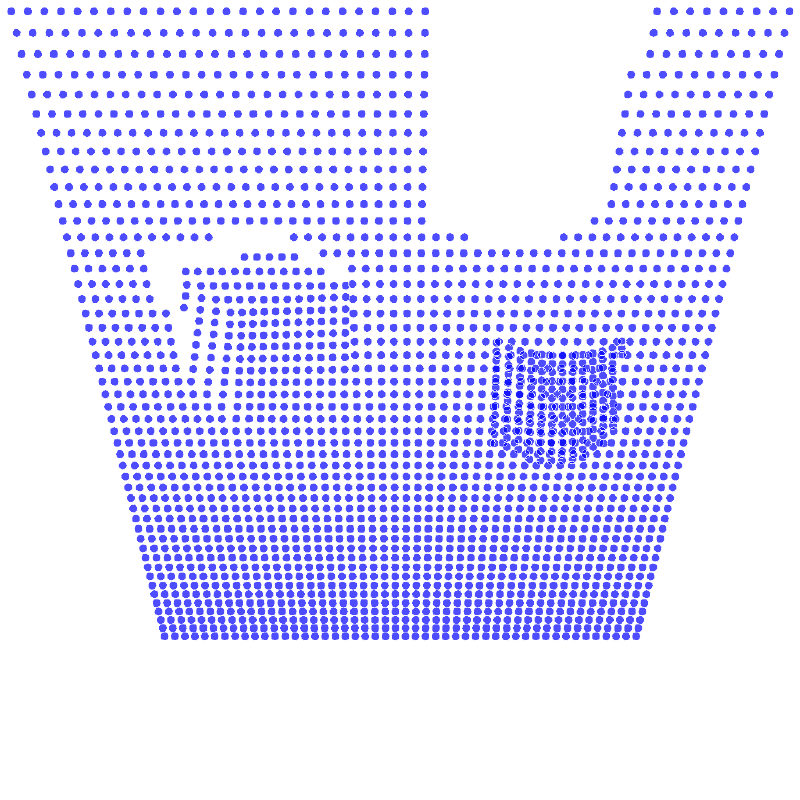}
\end{minipage}
\\
$P_2({\color{magenta}t_1},\pi_1)$%
\begin{minipage}[c]{5em}
\includegraphics[width=\textwidth,clip,trim={0 10em 0 0}]{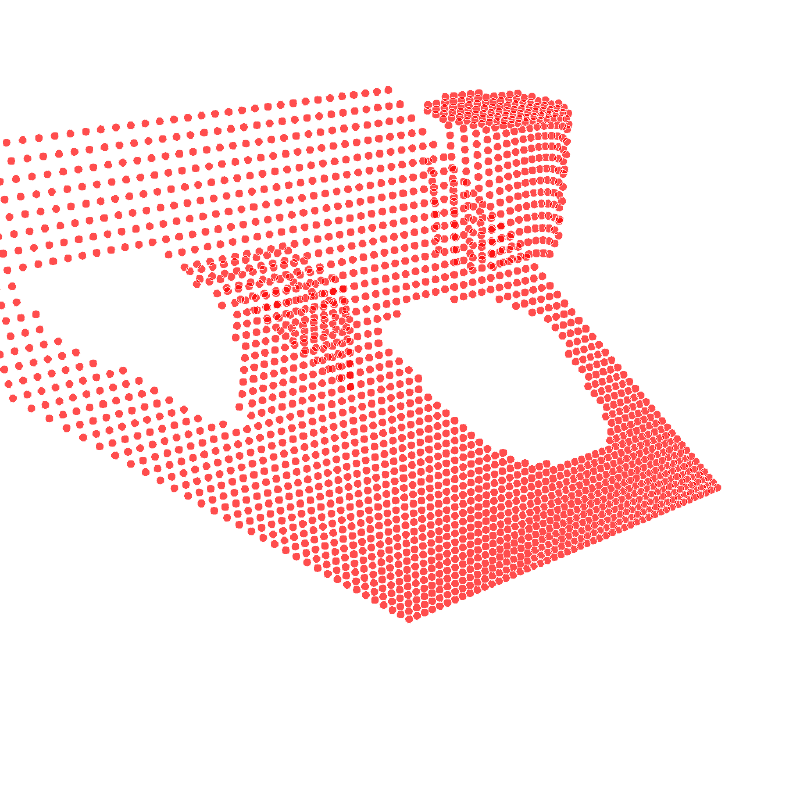}
\end{minipage}%
$P_2({\color{teal}t_2},\pi_1)$%
\begin{minipage}[c]{5em}\centering
\includegraphics[width=\textwidth,clip,trim={0 10em 0 0}]{Figures/P4_t4.pdf}
\end{minipage}
\end{minipage}%
~$\Rightarrow$~%
\begin{minipage}[c]{10em}\centering
\includegraphics[width=10em,clip,trim={0 10em 0 0}]{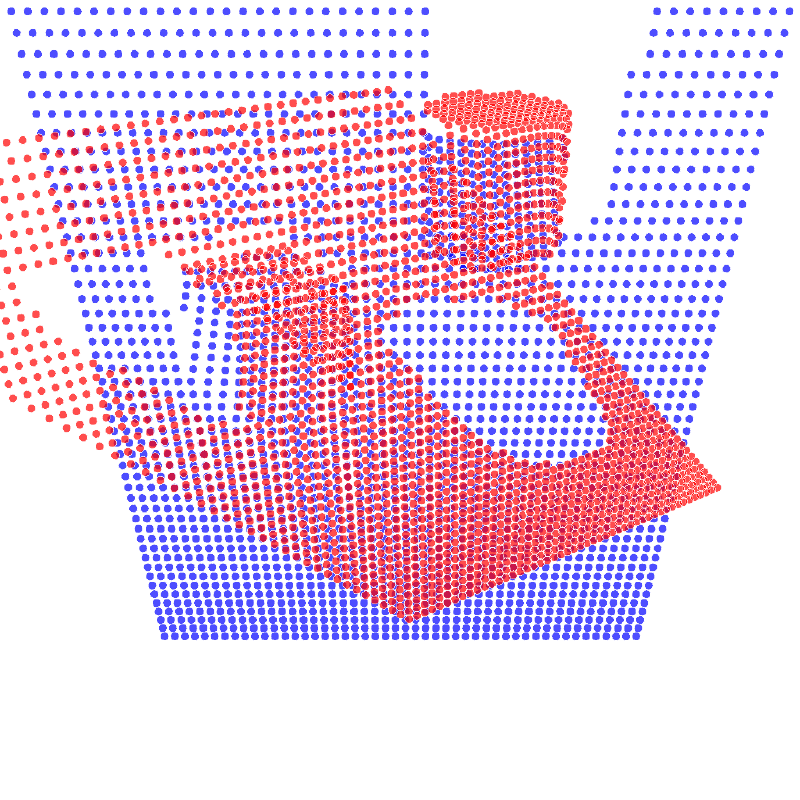}
${\color{magenta}t_1}$
\end{minipage}%
\begin{minipage}[c]{10em}\centering
\includegraphics[width=10em,clip,trim={0 10em 0 0}]{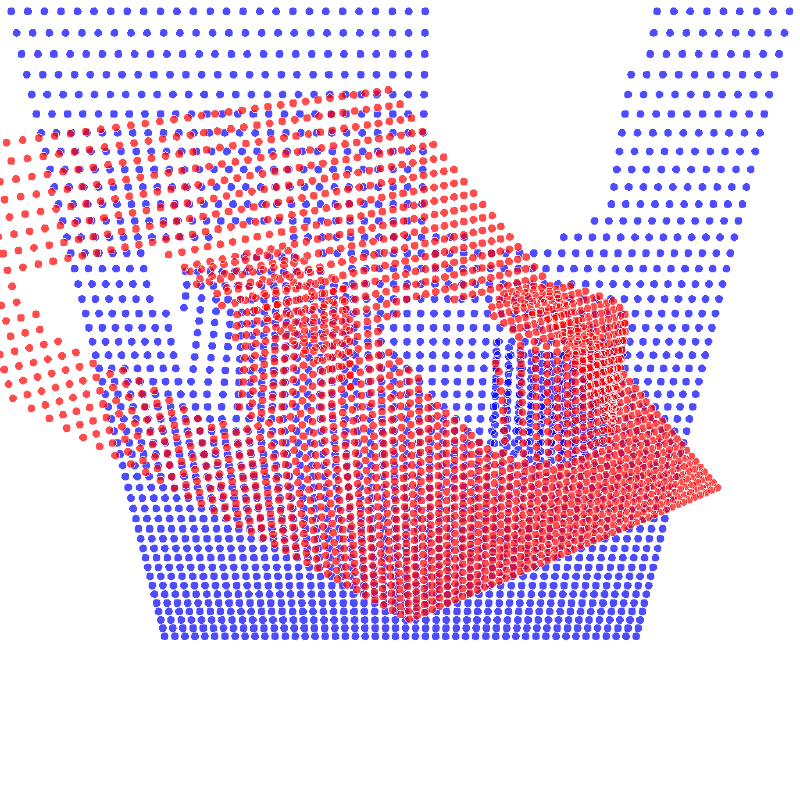}
${\color{teal}t_2}$
\end{minipage}
\vspace{-1em}
\caption{Left: Standard point maps applied to dynamic scenes as in \monster~\cite{zhang24monst3r:} fail to represent dynamics.
The cylinder, which is moving downwards, breaks invariance when the point maps are overlaid.
Right: Our \longmethod correctly represent dynamics by also controlling time in addition to viewpoint.
They allow to restore invariance while still representing the motion of the cylinder.
}%
\label{f:point-maps-comparison}
\end{figure*}

The impact of machine learning in 3D computer vision has been growing steadily.
For instance, \duster~\cite{wang24dust3r:}, a recent breakthrough, has proposed to learn a neural network that, given two images of a scene, maps each pixel to its corresponding 3D point, expressed in a shared 3D reference frame.
Notably, they show that knowledge of these viewpoint-invariant \emph{point maps} allows one to solve a variety of core 3D tasks such as estimating the camera intrinsics and extrinsics (by aligning pixels to their 3D points along camera rays), monocular depth estimation (by providing two identical copies of the same image to the model), and 2D matching (by comparing the reconstructed 3D points).
However, a key limitation is that their point maps cannot \emph{explain dynamic 3D contents}.

Dynamic scenes are ubiquitous in the real world, and interpreting and reconstructing them in 3D is potentially one of the most impactful applications of 3D computer vision, but also one of the most challenging.
Even state-of-the-art methods for dynamic 3D reconstruction~\cite{plizzari24spatial,stearns24dynamic,lei24mosca:} still use ad hoc designs, combining many different learned modules, including depth estimators, matchers, and segmenters, and require expensive and fragile test-time optimization.
This motivates us to consider how the simple and elegant approach of \duster could be extended to dynamic data.

The key question we aim to answer in this paper, then, is if and how the point maps representation could be used to tackle dynamic 3D reconstruction tasks.
The key to point maps is their invariance to the camera parameters, including viewpoint.
When the scene is static, only the camera changes, so viewpoint invariance is sufficient.
However, when the scene is dynamic, the 3D points themselves change over time.
We can still compute the point maps as defined in \duster; however, in this case, the outputs are \emph{not} invariant because, even if we fix the camera, the 3D points still move.

For example, this approach was considered by the recent \monster~\cite{zhang24monst3r:}.
While they obtain outstanding results, the technical limitations of their chosen representation are apparent.
Specifically, due to lack of invariance, they cannot predict corresponding 3D points directly, but need to combine their output with optical flow to do so.
We argue that easily establishing multi-view correspondences is one of the fundamental strengths of \duster's representation, and one that must be preserved in any extension to dynamic scenes.

In this paper, we introduce \emph{\longmethod} (\method), a new formulation that satisfies this requirement and that can easily extend \duster to dynamic scenes (\cref{f:model}).
Key to \method is the realization that invariance in dynamic scenes requires fixing both camera viewpoint and scene time.
Consider a pair of images of a scene, each of which defines a certain viewpoint and a timestamp.
We introduce for each image a \emph{pair of point maps} (\cref{f:point-maps-comparison}) sending each pixel to two versions of the corresponding `physical' 3D point, one corresponding to the timestamp of the first image and one to the timestamp of the second one.
Like in \duster, all 3D points are referred to the reference frame of the first image.

We argue that this is the minimal design that can tackle 4D tasks in full generality.
For static scenes, this is a direct generalization of \duster, as the two dual point maps are identical; it also generalizes \monster, as two of the four dual point maps match theirs.
Crucially, however, the dual point maps encode the scene \emph{motion} as well as \emph{dynamic correspondence}, eschewing the need to further process the data using, \eg, optical flow as done in \monster.
Specifically, because each physical point is reconstructed at both times, we can immediately infer scene flow and motion segmentation, and use this information to track the motion of rigid objects.
Furthermore, because physical points from different cameras are available in the same spatio-temporal reference frame, undoing the effect of viewpoint change (by fixing the viewpoint) and of scene motion (by fixing time), it is trivial to match them across views.
The latter also allows us to fuse point clouds despite motion in the scene.

We show empirically that the learned model can tackle dynamic scenes successfully, addressing the tasks discussed above.
In summary, our contributions are:
(1) To introduce the new concept of \longmethod which extends point maps to dynamic scenes in a way that allows solving many useful 3D and 4D reconstruction tasks;
(2) To show that the \duster model can be extended to output such dynamic maps, fine-tuning it on a mixture of datasets while generalizing well to real data;
(3) To demonstrate empirically the power of the resulting method in a number of tasks, from motion estimation to 4D reconstruction and rigid object tracking.
Overall, our results show that DPMs are very promising and can be a basis for new designs of 3D foundation models that can tackle dynamic scenes.


\section{Related Work}%
\label{sec:related}


\paragraph{Structure from Motion (SfM).}

Recovering the geometry of a static scene from a collection of images has been a long standing problem in computer vision~\cite{faugeras87motion,hartley00multiple,schaffalitzky02multi-view}.
COLMAP~\cite{schonberger16structure-from-motion} is perhaps the most popular implementation of the `classical' pipelines based on keypoint detection and description, matching, and bundle adjustment~\cite{agarwal09building,furukawa10towards}.
With the advent of machine learning, many authors have attempted to use neural networks for SfM, often to improve individual components like point matching~\cite{daniel17superpoint:,tyszkiewicz20disk:,yi16lift:,jiang21cotr:,chen21learning,lindenberger23lightglue:,sarlin20superglue:,shi22clustergnn:}, but also experimenting with fully-differentiable SfM pipelines~\cite{teed20deepv2d:,teed21droid-slam:,ummenhofer17demon:,wang21deep,wei20deepsfm:}.
VGGSfM~\cite{wang24vggsfm:} is perhaps as yey the most successful example of a method that can improve bundle adjustment by learning data priors and using them to address ambiguous situations such as matching texture-less regions.
Of particular relevance to this paper is \duster~\cite{wang24dust3r:}, which introduces a unified representation for predicting camera motion and 3D geometry jointly.

\paragraph{Non-rigid Structure from Motion (NR-SfM).}

In the case where there are dynamic elements in the scene the problem becomes more challenging because of much higher ambiguity, without the ability to directly triangulate points. 
Early works~\cite{bregler00recovering, torresani04learning} proposed assumptions under which the problem is meaningfully solvable.
Such assumptions were further refined in follow up works such as~\cite{valmadre12general,
akhter08nonrigid,
akhter11trajectory,
chen11trajectory}.
Several authors also considered dense dynamic reconstruction~\cite{kumar17monocular, ranftl16dense,russell14video}.

With the advent of neural radiance fields, several authors have considered extension of the classic NR-SfM problem where one reconstructs a model of the scene appearance as well.
Methods like
DyniBar~\cite{li23dynibar:},
DCT-NeRF~\cite{wang21neural},
NSFF~\cite{li21neural} and
Dynamic View Synthesis~\cite{gao21dynamic}
fit deformations between frames, whereas works such as
D-NeRF~\cite{pumarola21d-nerf:},
NeRFlow~\cite{du21neural},
Nerfies~\cite{park21nerfies:},
Space-time NeRF~\cite{xian21space-time},
Hypernerf~\cite{park21hypernerf:},
Deformable 3D Gaussians~\cite{yang24deformable},
4D GS~\cite{wu234d-gaussian} and the work of~\cite{yang24real-time}
fit instead deformations of each frame back to a shared canonical reconstruction.
Shape of Motion~\cite{wang24shape} models deformation longer term by trying to explicitly fit individual 3D Gaussian tracks to whole videos.
All these methods require expensive test-time optimization.

Concurrently to our work, \duster was recently extended by \monster~\cite{zhang24monst3r:}, CUT3R~\cite{wang2025continuous}, and Stereo4D~\cite{jin2024stereo4d} to tackle dynamic content with very good results.
However, \monster and CUT3R do not explore the full formulation design space and their extension only consider freezing time which cannot solve all the 4D tasks that our method can. Stereo4D does consider dynamic flow, but lacks the concept of invariant point clouds and the exploration of downstream tasks.

\paragraph{Optical Flow.}

Finding dense correspondences between images is usually cast as optical flow prediction~\cite{horn93determining,lucas81an-iterative,bay06surf:,brox09large,brox04high,bruhn05lucas/kanade}.
Early optical flow methods that used deep learning include
Flow Fields~\cite{bailer15flow},
FlowNet~\cite{dosovitskiy15flownet:,ilg16flownet},
PWC-Net~\cite{sun18pwc-net:} and
RAFT~\cite{teed20raft:}.
Some works~\cite{janai18unsupervised,ren18a-simple,shi23videoflow:} consider multi-frame optical flow
Multi-frame~\cite{janai18unsupervised}
Some~\cite{jiang21learning} suggested that the prior learned by neural networks is helpful to track points behind occlusions.
GMFLow~\cite{xu22gmflow:} suggest to consider optical flow estimation a global matching problem rather than a local one.
Others like FlowFormer~\cite{huang22flowformer:,shi23flowformer:} used transformer-based neural networks.

\paragraph{Scene Flow.}

The \emph{scene flow} is a set of 3D correspondences between scene points at two different times.
Examples of methods that can recover scene flow include RAFT-3D~\cite{teed21raft-3d:} and the very recent SpatialTracker~\cite{xiao24spatialtracker:} and SceneTracker~\cite{wang24scenetracker:}.
Shape of Motion~\cite{wang24shape} also estimates scene flow as a byproduct.
TAPVid-3D~\cite{koppula24tapvid-3d:} proposes a new benchmark to assess scene flow recovery.











\section{Method}%
\label{sec:method}

\newcommand{\p}{\boldsymbol{p}}
\newcommand{\bt}{\boldsymbol{t}}
\newcommand{\bu}{\boldsymbol{u}}

\begin{figure}[t]
\centering
\begin{minipage}[c]{0.275\textwidth}
\includegraphics[width=\textwidth]{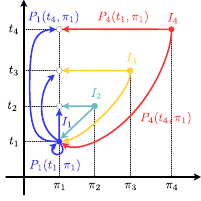}
\end{minipage}~%
\begin{minipage}[c]{0.1\textwidth}
\includegraphics[width=5em,trim={1em 8.4em 1em 0},clip]{Figures/P1_t4_P4_t4}
\includegraphics[width=5em,trim={1em 8.4em 1em 0},clip]{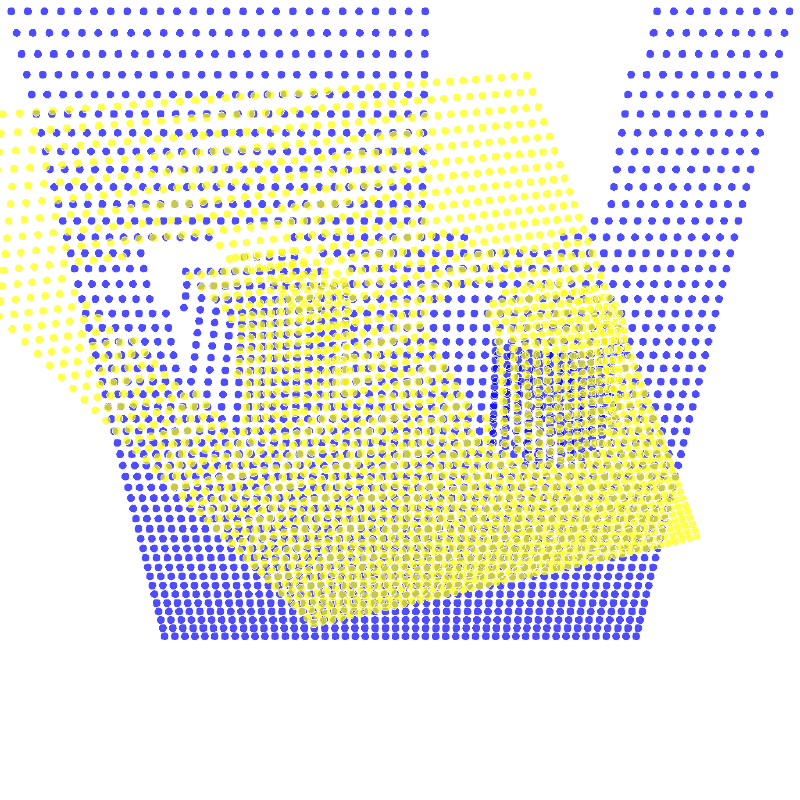}
\includegraphics[width=5em,trim={1em 8.4em 1em 0},clip]{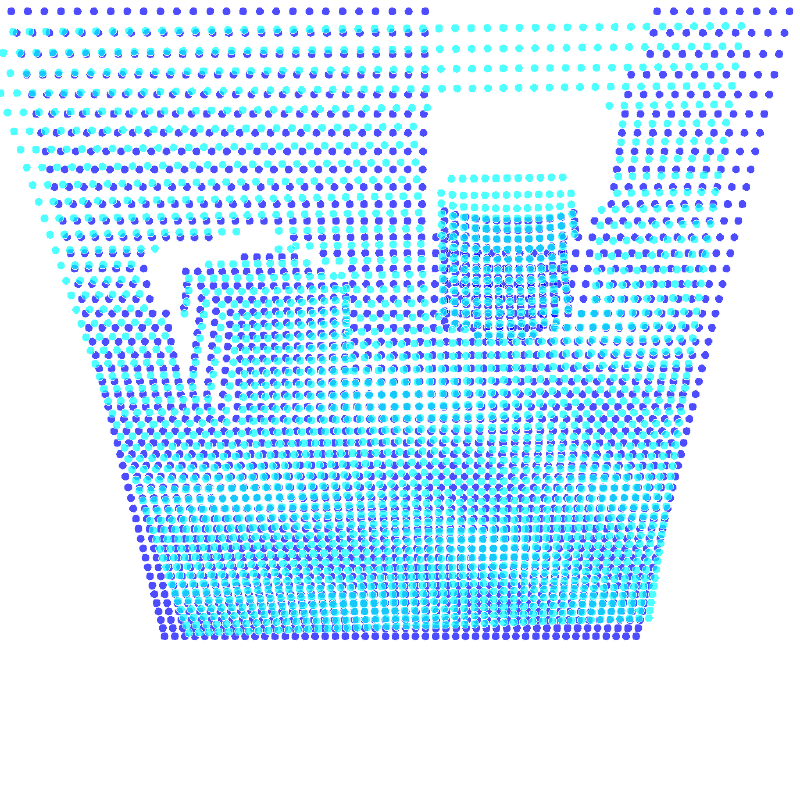}
\includegraphics[width=5em,trim={1em 8.4em 1em 0},clip]{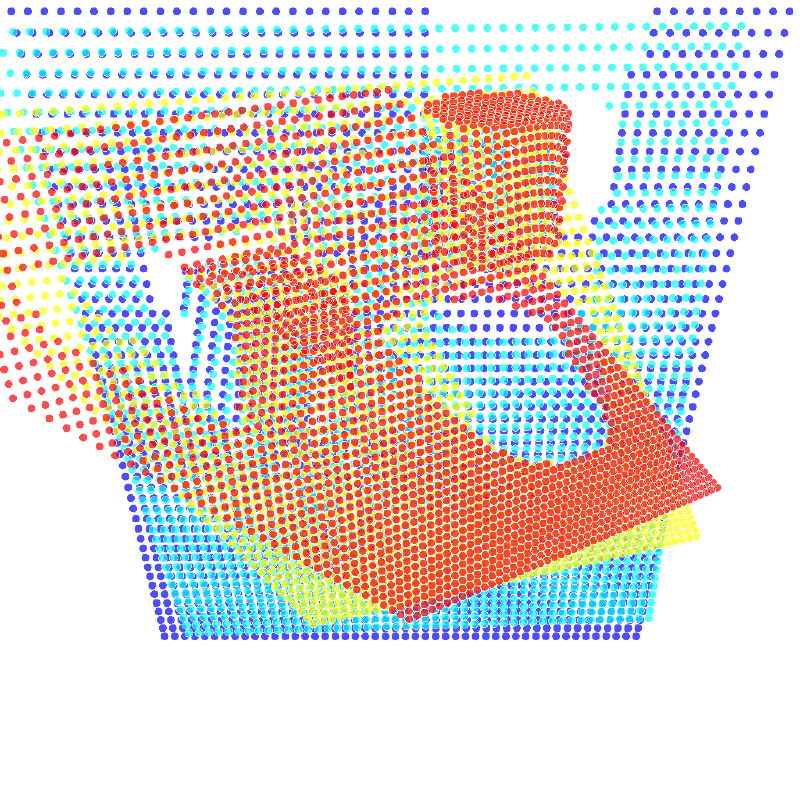}
\end{minipage}~%
\begin{minipage}[c]{0.1\textwidth}
\includegraphics[width=5em,trim={0em 4.5em 0em 0},clip]{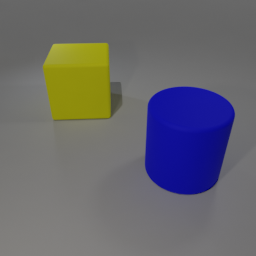}
\includegraphics[width=5em,trim={0em 4.5em 0em 0},clip]{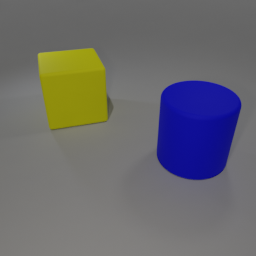}
\includegraphics[width=5em,trim={0em 4.5em 0em 0},clip]{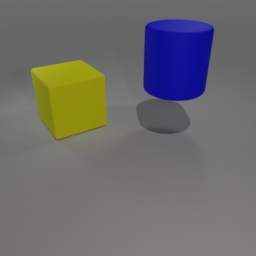}
\includegraphics[width=5em,trim={0em 4.5em 0em 0},clip]{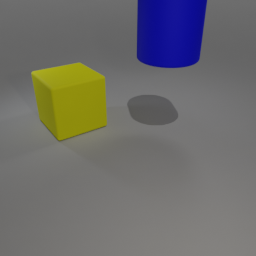}
\end{minipage}
\caption{Left: A schematic representation of the point cloud predictors $P_i(t_j,\pi_1)$ extracted from four images $I_1,\dots,I_4$ color coded as before.
Each circle represent an image, visualized as a point on a fictitious viewpoint-time plane.
Each arrow corresponds to a point map, the base of which is the source image and the tip of which is the reference viewpoint and time.
Right:
Visualization of the eight predicted point maps.
From bottom to top, we reconstruct the animated point cloud in the reference frame of the first image $\pi_1$.
In the bottom-right corner, each image contributes an invariant point map by undoing both the viewpoint and time changes; these point maps can then be fused. 
}%
\label{f:video-scheme}
\end{figure}

We start by reviewing the concept of point maps as used in \duster in \cref{sec:point-maps}.
We then introduce our new \longmethod representation in \cref{sec:dynamic-point-maps} and explain how we extend \duster to predict it. In \cref{sec:training-formulation} we describe how we train our model and discuss the training datasets.

\subsection{Point maps}%
\label{sec:point-maps}

Let $I$ be an image containing a scene we wish to reconstruct in 3D.
We cast this as the problem of predicting the corresponding \emph{point map} $P$ which associates each pixel $\bu$ with the corresponding scene point $\p = P(\bu) \in \mathbb{R}^3$, expressed in the reference frame $\pi$ of the camera.
A point map is thus similar to a depth map, but contains strictly more information than the latter.
In fact, the point map can be recovered from the depth map only if the intrinsic parameters of the camera (e.g., the focal length) are given; or, knowledge of the point map allows us to infer the camera intrinsics.

Next, consider two images $I_1$ and $I_2$ taken by two cameras with different viewpoints $\pi_1$ and $\pi_2$ and corresponding point maps $P_1$ and $P_2$.
\duster has recently proposed a model where both point maps are expressed in the same reference frame as the first image.
We make this explicit by writing $P_1(\pi_1)$ and $P_2(\pi_1)$, so that $P_i(\pi_j)$ means that the point map is extracted from image $i$ and expressed in the reference frame of camera $j$.
This simple change has profound consequences as it means that the point maps are \emph{viewpoint invariant}.
This means that, if $\bu_1$ and $\bu_2$ are two pixels that correspond to the same 3D point, then $P_1(\pi_1)(\bu_1) = P_2(\pi_1)(\bu_2)$.

As \duster noted, this invariance is sufficient to solve a number of core 3D tasks, such as estimating the camera extrinsics (\ie, the relative camera motion), establishing point correspondences between images, aligning and fusing the point clouds, and so on.
As noted above, it also allows us to infer the camera intrinsics and depth.
Their idea, then, is to learn a neural network $\Phi$ that could predict the point maps from the two images, as $(P_1(\pi_1),P_2(\pi_1)) = \Phi(I_1,I_2)$.
The fact that the point maps are viewpoint invariant actually simplifies the job of the network, as it makes the task more similar to a labeling problem.

The most significant limitation of the \duster model is that it cannot handle dynamic scenes.
We tackle this issue in the next section.

\subsection{Dynamic point maps}%
\label{sec:dynamic-point-maps}

Consider two images $I_1$ and $I_2$ taken at two different times $t_1$ and $t_2$.
If we consider the point maps $P_1(\pi_1)$ and $P_2(\pi_1)$ defined in \cref{sec:point-maps}, which is also the approach taken by~\cite{zhang24monst3r:}, we have a problem: the representation is \emph{no longer invariant}.
In fact, while both point maps are defined with respect to the \emph{same} reference frame $\pi_1$, the 3D points themselves move over time, so in general
$P_1(\pi_1)(\bu_1) \not= P_2(\pi_1)(\bu_2)$.
This is why \monster relies on an additional image matching network to establish correspondences $(\bu_1,\bu_2)$ between pixels, necessary for estimating scene motion and other tasks.

We can restore the invariance by controlling not only for viewpoint, but also for time, estimating point maps $P_1(t_1,\pi_1)$ and $P_2(t_1,\pi_1)$.
With this notation, we mean that the 3D points in both point maps are referred to the same reference frame $\pi_1$ \emph{and} time $t_1$ as the first camera, effectively undoing both viewpoint change and scene motion.
In this way, we can re-establish the invariance property $P_1(t_1,\pi_1)(\bu_1) = P_2(t_1,\pi_1)(\bu_2)$.
On the other hand, by undoing scene motion, we lose the ability to estimate it, which is key for motion analysis.

Our solution, then, is to introduce the concept of \emph{\longmethod} (\method).
By this, we mean predicting for each image a \emph{pair} of point maps, expressing the points in the two time stamps $t_1$ and $t_2$.
With our notation, image $I_1$ is associated with the pair of point maps $P_1(t_1,\pi_1)$ and $P_1(t_2,\pi_1)$ and image $I_2$ with point maps $P_2(t_1,\pi_1)$ and $P_2(t_2,\pi_1)$.

Note that all point maps are expressed in the same reference frame $\pi_1$ of the first camera.\footnote{The problem is symmetric, so we can obtain all quantities referred to the second camera by simply swapping the inputs to the network.}
Pairs with the same argument share the same spatio-temporal reference frame and are thus invariant.\footnote{
\Ie, 
$P_1(t_1,\pi_1)(\bu_1)= P_2(t_1,\pi_1)(\bu_2)$ and 
$P_1(t_2,\pi_1)(\bu_1)= P_2(t_2,\pi_1)(\bu_2)$
for all pairs of corresponding pixels $(\bu_1,\bu_2)$.
}
For example, we can establish correspondences between the two images by matching the 3D points in $P_1(t_1,\pi_1)$ and $P_2(t_1,\pi_1)$.
At the same time, we can recover scene flow simply by taking the difference $P_1(t_2,\pi_1) - P_1(t_1,\pi_1)$.

Just like \duster is a powerful representation for static scenes, \method is a powerful representation for dynamic ones.
In addition to subsuming all tasks that can be addressed in the static case (e.g., estimation of camera intrinsics and extrinsics, 3D reconstruction, etc.), they can also solve a number of tasks that are specific to 4D reconstruction, such as deformation-invariant matching and scene flow estimation, and help with plenty more, such as estimating rigid body motion.
In \cref{sec:reductions} we show some of these reductions more formally.
Most importantly, as we show in the next section, we can equip \duster with \method relatively easily.


\begin{figure}[htb]
\centering
\includegraphics[width=.95\linewidth]{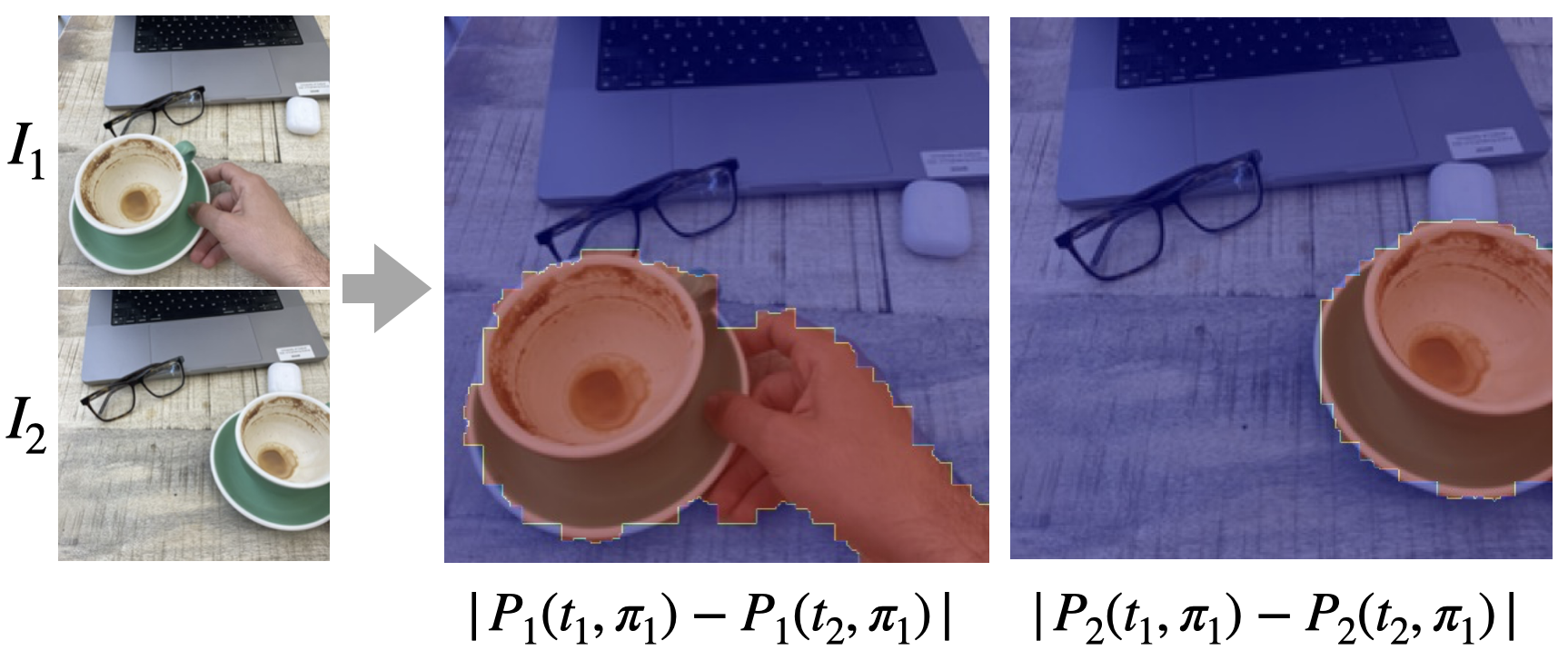}
\vspace{-1em}
\caption{\textbf{Motion segmentation:} from a pair of images using the dynamic point cloud predicted by the network we can segment out the dynamic elements of the scene, despite the camera motion.}%
\label{f:motion_seg}
\end{figure}

\paragraph{Dynamic \duster.}

Recall that \duster learns a network $\Phi$ that maps images $I_1$ and $I_2$ to a pair of point maps.
Here, we extend this network to simply output six instead of three channels per image by adding suitable heads, so that each image is mapped to two maps.
This can also be seen as a function
\begin{equation}\label{e:dynamic-duster}
\{P_i(t_j,\pi_1)\}_{i,j \in \{ 1, 2\}}
=
\Phi(I_1,I_2).
\end{equation}
Hence, four point maps are estimated in total, two for each image $i=1,2$ and time $j=1,2$.
Note that all point maps are referred to the reference frame $\pi_1$ of the first camera.\footnote{Switching to the second camera is trivial, as we can simply swap the inputs to the network, and recovers four more point maps, for a total of eight possible combinations.}
By predicting such a \method, this network is sufficient to solve a number of 4D tasks, as we have discussed above and further in \cref{sec:reductions}.



\subsection{Training formulation}%
\label{sec:training-formulation}

To supervise the model $\Phi$, we require video sequences with corresponding dynamic point maps; namely, the training data $\mathcal{D}$ is a collection of tuples
$
(
I_1, I_2,
P_1(t_1,\pi_1),
P_1(t_2,\pi_1),
P_2(t_1,\pi_1),
P_2(t_2,\pi_1)
).
$
To obtain such examples, we use a mixture of synthetic and real data providing various degrees of supervision. Synthetic data has already been shown to be very effective for a number of low-level computer vision tasks, such as optical flow~\cite{dosovitskiy15flownet:}, tracking~\cite{karaev24cotracker}, and depth prediction~\cite{yang24depthv2}.
With synthetic data, we have perfect knowledge of the underlying scene geometry, including its \emph{deformation} due to the dynamic parts of the scene.
With this, we can determine for each pixel $\bu$ in image $I_1$ which is the corresponding 3D point $\p(t_1,\pi_1)$.
Because we know the camera motion, we can then recover $\p(t_1,\pi_2)$.
Because we know the deformation of the scene, we can also recover $\p(t_2,\pi_1)$ and $\p(t_2,\pi_2)$ for image $I_2$.


\begin{figure}[htb]
\centering
\includegraphics[width=0.9\linewidth]{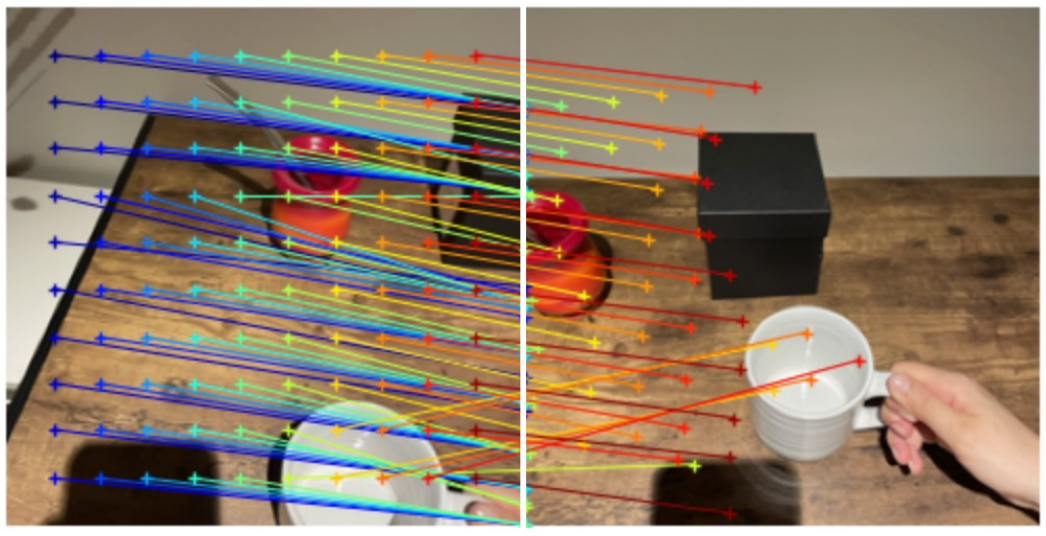}
\vspace{-1em}
\caption{\textbf{Point correspondence:} DPMs allow to recover correspondences accounting both for dynamic objects as well as camera motion even when challenging viewpoint changes.}%
\vspace{-.5em}
\label{f:matching}
\end{figure}

\paragraph{Training loss.}

We note that the scale of a 3D scene cannot be determined uniquely from any number of views; therefore, we relax the predictions to be determined up to a scaling factor.
In order to do so, let $P \in \mathbb{R}^{3 \times HW}$ be a ground-truth point cloud and let $\hat{P}$ be its predicted counterpart.
We then define per-pixel regression loss given by~\cite{wang24dust3r:}
$$
L_\text{reg}(\hat{P}, P, i) =
\left \|
\frac{
\hat{P}_{:,i}
}
{
 \frac{1}{HW} \sum_{j=1}^{HW} \| \hat{P}_{:,j} \|
}
-
\frac{
 P_{:,i}
}
{
 \frac{1}{HW} \sum_{j=1}^{HW} \| P_{:,j} \|
}
\right \|.
$$
This makes sense because points are expressed in the reference frame of one of the cameras, so, for example, $\|P_{:,i}\|$ is the distance of point $P_{:,i}$ with respect to the camera center. The full loss follows confidence calibrated formulation from ~\cite{wang24dust3r:,novotny17learning,kendall17what}:
\vspace{-2mm}
\begin{equation}
  L_\text{conf}(\hat{P}, P) = \frac{1}{HW} \sum_{i=1}^{HW} 
            C_i L_\text{reg}(\hat{P}, P, i) - \alpha \log C_i.
\label{eq:conf_loss}
\vspace{-2mm}
\end{equation}

Recall that network $\Phi$ predicts four separate point clouds.
For simplicity, we stack them into point clouds $P$ and $\hat{P}$ and minimize $L_\text{conf}(\hat{P}, P)$.
Again, this makes sense because all point clouds are defined in the same reference frame $\pi_1$.

\paragraph{Dataset details.}

We train our model on a total of 7 datasets (see \Cref{tab:datasets}).
We use the Kubric data pipeline~\cite{greff22kubric:} to generate a new synthetic dataset, which we call MOVi-G, to train our model following the Multi-Object Video (MOVi) dataset format but with more complex camera trajectories through spline interpolation and randomised intrinsics.
For generating dynamic point cloud ground truth, we use GT bounding box tracking and NOCS~\cite{wang19normalized} coordinates to transform points between different reference times and viewpoints.
We render 9,750 clips and sample 6 random pairs of images from each clip for a total of 58,500 training samples. We also add Waymo dataset~\cite{sun2020waymo} where we compute dynamic point maps from LiDAR data in a similar way.
For the rest of the datasets,  we either omit supervision for cross-time reconstructions ($P_2(t_1, \pi_1)$ and $P_1(t_2, \pi_1)$) or in the case of fully static scenes—they represent the identical geometry and therefore require no dynamic deformations.



\paragraph{Network details.}

The model architecture is based on the network design from \duster (\cref{f:model}).
For $T=2$, it shares the same backbone but adds two additional heads $\phi_{ij}$ for prediction, for a total of four regression outputs.
Each output comprises both a point map $P_i(t_j,\pi_1)$ and a confidence map $C_i(t_j,\pi_1) \in [0,1]^{HW}$, similar to~\cite{wang24dust3r:}.
They are thus predicted as
$
(P_i(t_j, \pi_1), C_i(t_j, \pi_1)) = \phi_{ij} (F)
$,
$
i,j \in \{1, 2\}
$,
where $F$ are the features computed by the transformer encoder-decoder backbone.
We initialize both heads associated with each point cloud with the single head from \duster, which should approximate at the start of training the static reconstruction component of that image.
We train our network with $(512, 288)$ and $(512, 336)$ resolutions.

\section{Evaluation}%
\label{sec:evaluation}

\begin{table*}
\centering
\label{tab:model_comparison}
\small
\tablestyle{8pt}{1.0}
\begin{tabular}{l cc cc cc cc cc}
\toprule
\multirow{2}{*}{\textbf{Model}} & \multicolumn{2}{c}{\textbf{Sintel}} & \multicolumn{2}{c}{\textbf{Point Odyssey}} & \multicolumn{2}{c}{\textbf{Bonn}} & \multicolumn{2}{c}{\textbf{Kubric}} & \multicolumn{2}{c}{\textbf{KITTI (crop)}} \\
\cmidrule(lr){2-3} \cmidrule(lr){4-5} \cmidrule(lr){6-7} \cmidrule(lr){8-9} \cmidrule(lr){10-11}
 & Abs Rel & $\delta < 1.25$ & Abs Rel & $\delta < 1.25$ & Abs Rel & $\delta < 1.25$ & Abs Rel & $\delta < 1.25$ & Abs Rel & $\delta < 1.25$ \\
\midrule
MonST3R & 0.347 & \textbf{0.573} & 0.065 & 0.953 & \textbf{0.071} & \textbf{0.941} & 0.166 & 0.779 & 0.069 & 0.945 \\
DPM & \textbf{0.321} & 0.564 & \textbf{0.059} & \textbf{0.957} & 0.082 & 0.919 & \textbf{0.078} & \textbf{0.950} & \textbf{0.052} & \textbf{0.968} \\
\bottomrule
\end{tabular}
\vspace{-1em}
\caption{\textbf{Depth Evaluation from 2-View Input}: Comparison of MonST3R and DPM across Sintel, Point Odyssey, Bonn, Kubric, and KITTI datasets. Our DPM model consistently yields lower absolute relative errors on challenging benchmarks—delivering an average reduction of roughly 17.5\% in absolute relative error.}
\vspace{-1em}

\label{tab:2_view_depth}
\end{table*}

\begin{table*}[htbp]
\centering
\small 
\tablestyle{6pt}{1.0}
\begin{tabular}{lc cc cc cc cc}
\toprule
\multirow{2}{*}{Category} & \multirow{2}{*}{Method} & \multicolumn{2}{c}{\textbf{Sintel}} & \multicolumn{2}{c}{\textbf{Bonn}} & \multicolumn{2}{c}{\textbf{KITTI}} & \multicolumn{2}{c}{\textbf{KITTI (crop)}} \\
\cmidrule(lr){3-4} \cmidrule(lr){5-6} \cmidrule(lr){7-8} \cmidrule(lr){9-10}
 &  & Abs Rel $\downarrow$ & $\delta<1.25$ $\uparrow$ & Abs Rel $\downarrow$ & $\delta<1.25$ $\uparrow$ & Abs Rel $\downarrow$ & $\delta<1.25$ $\uparrow$ & Abs Rel $\downarrow$ & $\delta<1.25$ $\uparrow$ \\
\midrule
\multirow{2}{*}{\footnotesize 1-frame} & Marigold & 0.532 & 51.5 & \textbf{0.091} & \textbf{93.1} & 0.149 & 79.6 & - & - \\
 & DepthAnythingV2 & \textbf{0.367} & \textbf{55.4} & 0.106 & 92.1 & \textbf{0.140} & \textbf{80.4} & - & - \\
\cmidrule(lr){1-10}
\multirow{3}{*}{\footnotesize Video depth} & NVDS & 0.408 & 48.3 & 0.167 & 76.6 & 0.253 & 58.8 & - & - \\
 & ChronoDepth & 0.687 & 48.6 & 0.100 & 91.1 & 0.167 & 75.9 & - & - \\
 & DepthCrafter & \textbf{0.292} & \textbf{69.7} & \textbf{0.075} & \textbf{97.1} & \textbf{0.110} & \textbf{88.1} & - & - \\
\cmidrule(lr){1-10}
\multirow{4}{*}{\footnotesize Joint D\&P} & Robust-CVD & 0.703 & 47.8 & - & - & - & - & - & - \\
 & CasualSAM & 0.387 & 54.7 & 0.169 & 73.7 & 0.246 & 62.2 & - & - \\
 & MonST3R & 0.335 & \textbf{58.5} & \textbf{0.063} & \textbf{96.4} & \textbf{0.104} & \textbf{89.5} & 0.111 & 87.2 \\
 & DPM & \textbf{0.328} & 54.6 & 0.068 & 93.9 & 0.140 & 78.2 & \textbf{0.097} & \textbf{89.1} \\
\bottomrule
\end{tabular}
\vspace{-1em}

\caption{\textbf{Video Depth Evaluation:} Performance Comparison between single-frame, video depth, and optimization-based models on Sintel, Bonn, and KITTI datasets. Our approach demonstrates robust performance across all datasets.}
\vspace{-1em}

\label{tab:depth_estimation_comparison}
\end{table*}

\begin{table}[t]
\centering
\tablestyle{1pt}{1.0}
\resizebox{\linewidth}{!}{
\begin{tabular}{ll cccc cc}
\toprule
 & & \multicolumn{4}{c}{$L_\text{rel}$} & \multicolumn{2}{c}{\textbf{Object Pose Error}} \\
\cmidrule(lr){3-6} \cmidrule(lr){7-8}
Dataset & Method & $P_1(t_1)$ & $P_2(t_1)$ & $P_1(t_2)$ & $P_2(t_2)$ & RPE rot & RPE trans \\
\midrule
\multirow{2}{*}{Kub.-F} & MonST3R & 0.209 & 0.275 & 0.394 & 0.201 & 56.1 & 0.504 \\
 & Ours & \textbf{0.041} & \textbf{0.047} & \textbf{0.049} & \textbf{0.035} & \textbf{33.7} & \textbf{0.053} \\
\midrule
\multirow{2}{*}{Kub.-G} & MonST3R & 0.163 & 0.265 & 0.346 & 0.178 & \multicolumn{2}{c}{\multirow{2}{*}{\textcolor{gray}{N/A}}} \\
 & Ours & \textbf{0.057} & \textbf{0.071} & \textbf{0.079} & \textbf{0.058} & \multicolumn{2}{c}{} \\
\midrule
\multirow{2}{*}{Waymo} & MonST3R & 0.197 & 0.221 & 0.249 & 0.178 & \multicolumn{2}{c}{\multirow{2}{*}{\textcolor{gray}{N/A}}} \\
 & Ours & \textbf{0.068} & \textbf{0.065} & \textbf{0.067} & \textbf{0.065} & \multicolumn{2}{c}{} \\
\bottomrule
\end{tabular}
}
\vspace{-1em}
\caption{\textbf{Dynamic reconstruction}. We compare our method with \monster~\cite{zhang24monst3r:}+RAFT~\cite{teed20raft:} on two tasks: relative point cloud error and object pose tracking. \monster struggles with predicting motion whereas our method explicitly trained on this task consistently performs better across datasets and metrics. $P_k(t_k)$ denotes $P_k(t_k, \pi_1)$ for brevity.}%

\vspace{-1em}

\label{tab:dynamic}
\end{table}

\begin{table}[h]
    \centering
    \tablestyle{1pt}{1.0}
    \small
    \begin{tabular}{l l l c c c c}
        \toprule
        & & & \multicolumn{2}{c}{\textbf{Scene Flow }} & \multicolumn{2}{c}{\textbf{Object Flow}} \\
        \cmidrule(lr){4-5} \cmidrule(lr){6-7}
        Dataset & Method & Input & Forward & Backward & Forward & Backward \\
        \midrule
        \multirow{3}{*}{Kub.-F} & MonST3R  & RGB  & 0.321 & 0.241 & 0.334 & 0.215 \\
                                   & RAFT-3D  & RGB\textcolor{red}{D}  & \textbf{0.051} & \textbf{0.054} & \textcolor{gray}{N/A} & \textcolor{gray}{N/A} \\
                                   & Ours     & RGB  & 0.081 & 0.071 & \textbf{0.033} & \textbf{0.029} \\
        \midrule
        \multirow{3}{*}{Kub.-G} & MonST3R  & RGB  & 0.334 & 0.279 & 0.310 & 0.265 \\
                                   & RAFT-3D  & RGB\textcolor{red}{D}  & 4.067 & 4.084 & \textcolor{gray}{N/A} & \textcolor{gray}{N/A} \\
                                   & Ours     & RGB  & \textbf{0.104} & \textbf{0.106} & \textbf{0.059} & \textbf{0.050} \\
        \midrule
        \multirow{3}{*}{Waymo} & MonST3R  & RGB  & 0.161 & 0.135 & 0.108 & 0.102 \\
                               & RAFT-3D  & RGB\textcolor{red}{D}  & 0.150 & 0.145 & \textcolor{gray}{N/A} & \textcolor{gray}{N/A} \\
                               & Ours     & RGB  & \textbf{0.051} & \textbf{0.053} & \textbf{0.020} & \textbf{0.020} \\
        \bottomrule
    \end{tabular}
\vspace{-1em}

    \caption{\textbf{3D End-Point Error (EPE) for Scene Flow and Object Flow:} Despite relying solely on RGB input, our method performs comparably to RAFT-3D~\cite{teed20raft:}, which utilizes ground truth depth (\textcolor{red}{D}), in Kubric-F and surpasses it in Kubric-G and Waymo. For Object Flow estimation, RAFT-3D is incapable of performing the task, whereas our method achieves the best results across all datasets.}
\vspace{-1em}
    
    \label{tab:3d_epe_results}
\end{table}

We evaluate \method on several 3D and 4D reconstruction tasks.
In \cref{sec:depth_eval}, we show that \method is on par with the state-of-the-art in depth estimation.
Next, in \cref{sec:dynamic_recon}, we shift our focus to dynamic 3D reconstruction, our key novelty, and show that here \method is significantly better than alternatives.
Finally, in \cref{sec:real_data}, we provide a qualitative analysis.

\begin{figure}[t]
\centering
\includegraphics[width=\linewidth]{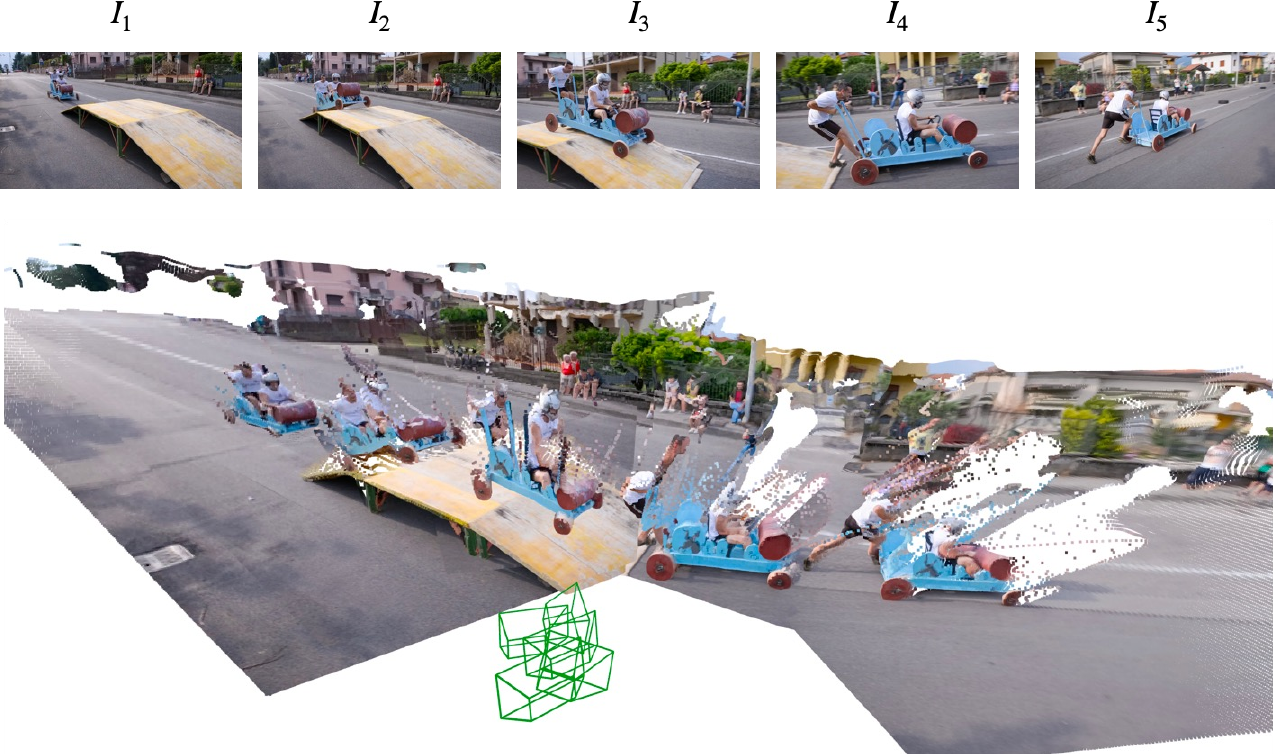}
\caption{\textbf{Camera tracking:} the explicit modeling of a dynamic point cloud allows to recover camera motion by directly aligning point clouds with matching timestamp.}%
\vspace{-1em}
\label{f:cams}
\end{figure}
\begin{figure*}
\centering
\includegraphics[width=\linewidth]{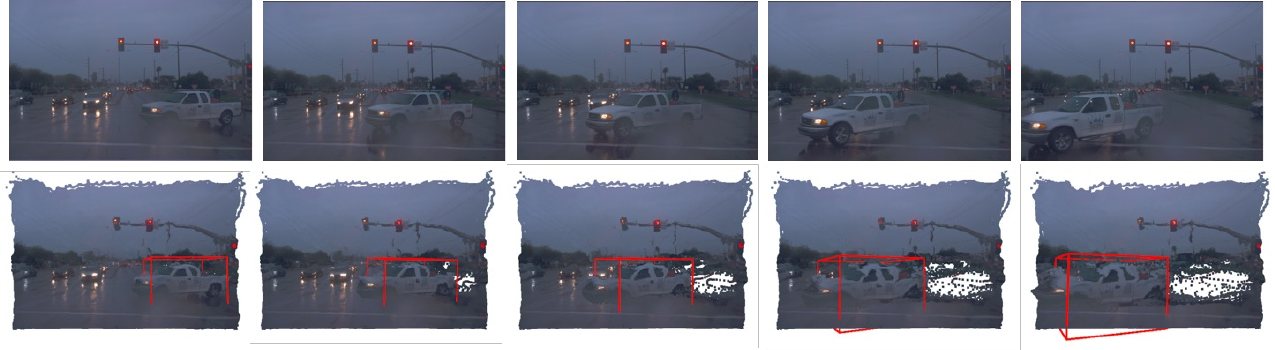}
\vspace{-2em}
\caption{\textbf{Rigid object motion:} 3D bounding box object tracking by masking and aligning the predicted dynamic point clouds.}%
\label{f:obj_track}
\end{figure*}
\begin{figure}[t]
\centering
\includegraphics[width=0.95\linewidth]{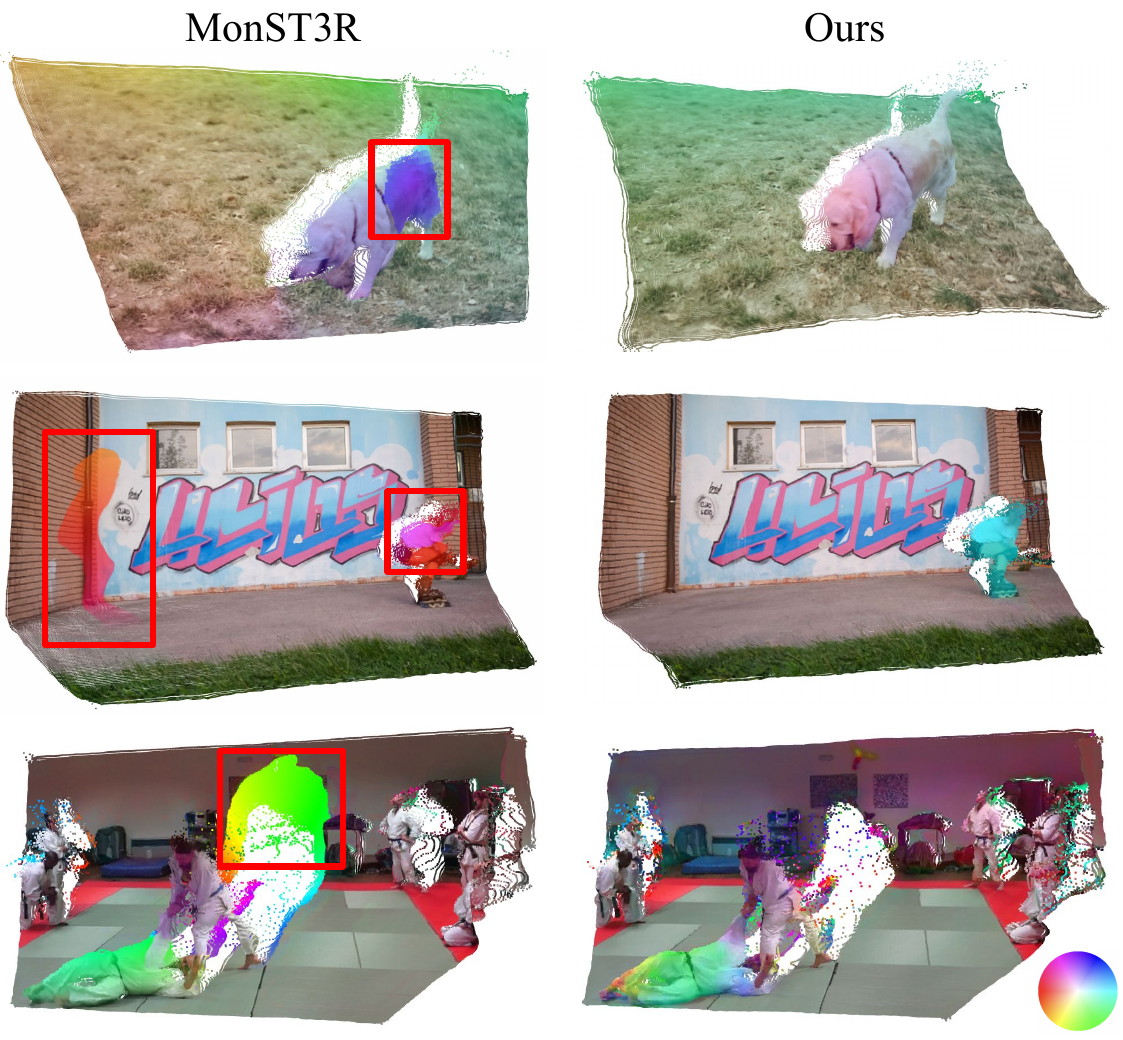}\vspace{-1em}
\caption{\textbf{Visualization of scene flow estimation results} for MonST3R (left) and our method (right). Red boxes highlight errors in MonST3R's predictions. \textbf{Row 1:} MonST3R exhibits unnatural artifacts and incorrect flow estimation. \textbf{Row 2:} Incorrect flow direction. \textbf{Row 3:} Poor handling of disocclusions due to MonST3R using 2D optical flow for warping. Our method consistently produces more accurate scene flow predictions across all cases.}
\label{f:sceneflow}
\end{figure}

\subsection{Depth prediction}%
\label{sec:depth_eval}

\paragraph{Two-view depth evaluation.}

Since \method is a generalization of \duster, we first evaluate it on the task of stereo depth prediction in \cref{tab:2_view_depth}.
We consider several standard benchmarks of dynamic scenes, including Bonn, Sintel, Point Odyssey, Kubric, and KITTI (crop).\footnote{We use a cropped version of KITTI because the original images have an extreme aspect ratio (512$\times$144).}
We report standard depth metrics: Absolute Relative Error (Abs Rel) and $\delta <1.25$ accuracy~\cite{yang24depth}.
Stereo pairs are formed by choosing pairs of frames at random within a set margin.
The results show that \method outperforms \monster on all datasets except Bonn.

\paragraph{Video depth evaluation.}

Next, in \cref{tab:depth_estimation_comparison}, we consider depth prediction for longer video sequences.
To fuse pairwise predictions from our network, we use a bundle adjustment strategy similar to that of MonST3R, but with a key modification: we remove the optical flow loss, as this is unnecessary when using the \method representation.
This eliminates the overhead associated with computing optical flow using an additional model, resulting in a more efficient pipeline while still delivering high-quality depth predictions.
We evaluate this approach on Sintel, Bonn, and KITTI datasets and show competitive performance with competitors including \monster.

\subsection{Dynamic reconstruction}%
\label{sec:dynamic_recon}

We evaluate our method on dynamic reconstruction using a test set consisting of 250 held-out clips from our MOVi-G dataset, as well as on the public Kubric MOVi-F dataset and Waymo Open.
These two variants of Kubric are fairly different:
MOVi-G has complex camera motion and a mix of static and dynamic objects, whereas MOVi-F has simple linear motion and also a mix of dynamic and static objects with varying amounts of motion blur.

We primarily compare against \monster~\cite{zhang24monst3r:}, which is trained with dynamic content but only predicts two of our four point maps (\cref{f:point-maps-comparison}).
To approximate the remaining point maps, \monster needs to use the pixel associations given by optical flow prediction, which is limited for pixels that are visible in both images.

\paragraph{Dynamic Point Maps prediction.}

We first show that we can predict our dynamic point maps well by using a fine-tuned version of \duster.
To this end, let $P, \hat P \in \mathbb{R}^{3\times HW}$ be, respectively, ground-truth and predicted maps.
Furthermore, let $M \subset \{1,\dots, HW\}$ be a mask representing a subset of image pixels.
Before computing the loss, we normalize the stacking $P$ of point maps $P_1(t_1,\pi_1)$ and $P_2(t_2,\pi_1)$ by the median 2-norm of its valid points $Z_{\text{med}} = \text{median}(\{\|P_{:,i}\|_2 \mid i \in M\})$.
We do the same for $\hat P$.
Then, the \emph{relative error} is
$$
L_\text{rel}(\hat P, P|M,s) = \frac{1}{|M|} \sum_{i \in M} 
\frac{\| \hat P_{:,i} - P_{:,i} \|}{\|P_{:,i}\|}.
$$
\Cref{tab:dynamic} report this metric for the four point map predictions.
Our method outperforms \monster both on synthetic Kubric and real-world Waymo datasets.
On the challenging Kubric-G dataset, the performance of \monster significantly degrades when making cross-time predictions ($P_2(t_1)$ and $P_1(t_2)$), highlighting the effectiveness of our representation for dynamic reconstruction.

\paragraph{Scene Flow.}

\method allows us to infer 3D point correspondences, or dense scene flow.
We assess this capability using standard metrics and protocols, further described in the~\cref{sec:appendix-scene-flow-metrics}.

\Cref{tab:3d_epe_results} presents the quantitative evaluation of 3D End-Point Error (EPE), defined as the average Euclidean distance between predicted and ground-truth 3D displacement vectors, for Scene Flow (overall 3D motion including camera movement) and Object Flow (object-only motion from a fixed viewpoint) Evaluations are conducted on the Kubric-F, Kubric-G, and Waymo datasets for three methods:
\monster~\cite{zhang24monst3r:} and RAFT-3D~\cite{teed21raft-3d:}, a scene flow method. 

Notably, RAFT-3D leverages ground truth depth (RGBD input), giving it an inherent advantage.
Despite this, our RGB-only method achieves comparable performance to RAFT-3D on Kubric-F and surpasses it significantly on Kubric-G and Waymo.
We further observe that RAFT-3D, being a specialized Scene Flow model, cannot perform Object Flow estimation—a closely related task—which highlights the greater flexibility and broader applicability of our method.
Compared to \monster, on average, our approach achieves approximately 76\% lower error across all datasets.
Although \monster shares structural similarities with our method, it lacks the capacity to explicitly capture temporal invariances, necessitating an additional optical flow model for temporal warping, thus resulting in increased errors.

\Cref{f:sceneflow} qualitatively demonstrates scene flow estimation, highlighting errors by \monster with red boxes.
Specifically, {Row 1} shows ghosting artifacts and incorrect motion estimations; {Row 2} illustrates incorrect flow direction predictions; and {Row 3} demonstrates \monster's poor handling of disocclusions due to reliance on 2D optical flow-based warping, which is inadequate for accurately capturing complex temporal dynamics.
These observations underscore our model's superior handling of challenging scenarios such as occlusions and disocclusions, thanks to its inherent temporal modeling.

\paragraph{Object tracking.}

We measure the performance of tracking of dynamic objects by estimating relative pose transformation between times $t_0$ and $t_1$.
We estimate the relative rotation $R$ and translation $t$ that align $P_1(t_1,\pi_1)$ to $P_1(t_2,\pi_1)$ using Umeyama algorithm~\cite{umeyama91least-squares}.
We obtain the reference transformations $\hat R$ and $\hat t$ from ground truth bounding boxes and compute the geodesic distance $\text{dist}(R, \hat R) = \arccos ((\text{tr}(R^{T} \hat R) - 1) / 2)$ and the $L_2$ distance between object centre translations $\text{dist}(t, \hat t) = \| t - \hat t\|_2$.

\Cref{tab:dynamic} shows that \monster fails to accurately predict future object locations because it lacks training for cross-time predictions and depends only on optical flow for correspondences. While rotation errors remain high for both methods during large object motions, our approach reduces these errors by 40\% compared to \monster.

\subsection{Qualitative results}%
\label{sec:real_data}

We test \method qualitatively in video clips recorded with a consumer camera by running the network on image pairs, please see also the supplemental.
\Cref{fig:teaser} shows a dynamic reconstructions for clips in which both an object and the camera are moving.
We observe that our method can disentangle the two motions, in particular by removing the camera motion.
\Cref{f:cams} visualises the recovered camera poses and dynamic point cloud on a DAVIS sequence.
We visualise the results of applying \method to solve downstream tasks in other examples, such as motion segmentation in \cref{f:motion_seg} and point correspondence in \cref{f:matching}.

\section{Downstream Applications}%
\label{sec:applications}

We now illustrate the ability of DPMs to easily solve a variety of 4D reconstruction tasks.
We refer the reader to~\cref{sec:reduction} for  formal derivations.

Firstly, we consider \emph{4D reconstruction}.
The point maps $P_i(t_i,\pi_1)$ provide a 3D reconstruction of each image $I_i$ in a sequence where the viewpoint is fixed to $\p_1$.
In this way, we obtain a a 4D `animation' of the scene where the intrinsic motion of the objects is isolated from the camera motion.
An example is shown in \cref{f:cams}.

Secondly, we consider \emph{motion segmentation}.
Give two (or more) images $I_1$ and $I_2$, we can simply compare point maps $P_1(t_1,\pi_1)$ and $P_1(t_2,\pi_2)$ to determine which 3D points from image $I_1$ have moved (independently of the camera motion).
The other two point maps allow us to do the same for image $I_2$.
By thresholding the difference between the two sets of points, as shown in \cref{f:motion_seg}.

Thirdly, we consider \emph{point correspondence}.
As noted, a key property of DPM is to restore invariance of the point maps despite in-scene camera motion which allows to match them.
Namely, given a pixel $\bu_1$ in image $I_1$, we can described it via the 3D point $\p_1 = P_1(t_1,\pi_1)(\bu_1)$; to find the matching point in image $I_2$, we then simply compare $\p_1$ to all points $\p_2(\bu_2)= P_1(t_2,\pi_2)(\bu_2)$, and take the pixel $\bu_2$ that minimize the distance $\|\p_2(\bu_2) - \p_1\|$ as the match.
An example of this process is shown in \cref{f:matching}.

Fourthly, we consider \emph{camera tracking}.
One way to recover the camera motion while ignoring dynamic distractors
is to match point clouds $P_1(t_1,\pi_1)$ and $P_2(t_1,\pi_2)$ using Procrustes analysis, where crucially the time $t_1$ is fixed, but the camera $\pi_i$ varies.
This effectively reduces to standard camera tracking with a static scene by undoing the passing of time.
An example is shown in \cref{f:cams}.

Lastly, we consider \emph{object tracking}.
In this case, given a mask $M$ for the object of interest in image $I_1$, one can simply observe points $M \odot P(t_i,\pi_1)$ to infer the motion of the object, independently of the camera motion.
If the object is rigid, it is easy to fit a rigid transformation on top.
An example is shown in \cref{f:obj_track}.


\section{Conclusions}%
\label{s:conclusions}

We have presented \longmethod, a new formulation that affords simple reduction of a number of difficult 4D reconstruction tasks to a single neural network.
This network densely predict 3D point clouds from a pair of images, controlling for both the spatial and time reference frames where the points are defined.
In this manner, it allows identifying congruent scene points despite camera and scene motion, as well as easily solving various 3D and 4D reconstruction tasks.
Numerous evaluations of our model on a series of benchmarks demonstrate that it consistently matches or surpasses state-of-the-art approaches across multiple tasks including video depth estimation, dynamic point cloud reconstruction and scene flow estimation.

We envision that our formulation will serve as a foundational framework for advancing future 3D/4D vision models. An especially promising extension involves integrating our approach into video models capable of jointly processing multiple frames, thereby allowing more accurate 4D reconstruction. Additionally, exploring unsupervised or semi-supervised training methods could further broaden its applicability and scalability.

\paragraph*{Acknowledgments.}

The authors of this work were supported by ERC 101001212-UNION.

{
\small
\bibliographystyle{ieeenat_fullname}
\bibliography{vedaldi_general,vedaldi_specific,main}
}
\appendix
\clearpage
\setcounter{page}{1}
\maketitlesupplementary{}

\section{Theory of \longmethod}%
\label{sec:reductions}

We discuss more formally how \longmethod are defined and how they can be used to solve various 4D reconstruction tasks.

\subsection{Monocular point maps}%
\label{sec:appendix-point-maps}

We represent the image $I$ as a $3 \times HW$ matrix by stacking the spatial dimensions, where 3 is the number of colour channels.
We assume that the image is taken by a pinhole camera.
Hence, a 3D point
$
\p=(x,y,z) \in \mathbb{R}^3
$
expressed in the reference frame of this camera projects to the image pixel
$
\bu=(u_x,u_y,1)
$
such that
$$
\bu \lambda =  K \p
$$
where $K\in\mathbb{R}^{3\times 3}$ is the camera's calibration matrix containing the camera's intrinsics parameters, and $\lambda > 0$ is the depth of the point.

The point map $P$ is just the collection of 3D point corresponding to each pixel, which we can write as a matrix $P \in \mathbb{R}^{3 \times HW}$.
Denote by $U \in \mathbb{R}^3$ the grid of pixels in homogeneous coordinates (this matrix is fixed).
Then we can write
$$
U \operatorname{diag} \Lambda = K P
$$
where $\Lambda \in \mathbb{R}_+^{HW}$ contains the depth values.

For monocular prediction, we can task a neural network $\Phi$ with mapping the image $I$ to the corresponding 3D point cloud $P$, i.e,
$
P = \Phi(I).
$
This problem is related to monocular depth estimation, in which a neural network $\Lambda = \Phi_\text{depth}(I)$ is tasked with associating pixels to depth values, but provides more information.
In fact, in order to reconstruct the 3D points $P$ from the depth $\Lambda$, we also require knowledge of the camera intrinsics $K$, so that the 3D points can be recovered as
$
P 
= K^{-1} U \operatorname{diag} (\Lambda)
= K^{-1} U \operatorname{diag} (\Phi_\text{depth}(I))
$.
On the contrary, knowledge of the point mao $P$ allow us to \emph{infer} depth and intrinsics by solving the equation
$
U \Lambda = K P = K \Phi(I)
$
in $\Lambda$ and $K$.

\subsection{Binocular point maps (\duster)}%
\label{sec:binocular-point-maps}

Next, we extend the case above to consider a pair of images $I_1$ and $I_2$.
Each image is taken by a camera with different intrinsics $K_1$ and $K_2$ and, most importantly, different viewpoints $\pi_1$ and $\pi_2$.

Let the symbols $\p(\pi_1)$ and $\p(\pi_2)$ be the coordinate of a certain 3D point $p$ expressed in the reference frames of the first and second cameras, respectively.
Following DUSt3R, we task a neural network $\Phi$ with predicting the pair of point clouds $(P_1(\pi_1),P_2(\pi_1))$ from the pair of images $(I_1,I_2)$:
\begin{equation}\label{e:duster}
(P_1(\pi_1),P_2(\pi_1)) =\Phi(I_1, I_2).
\end{equation}

As above, knowledge of $P_1(\pi_1)$ allows to recover the intrinsics $K_1$ of the first camera from
$
U_1 \Lambda_1 = K_1 P_1(\pi_1).
$
$K_2$ can be recovered as before by calling the network a second time with the two arguments swapped.
More interestingly, however, the second point cloud $P_2(\pi_1)$ contains the 3D point that corresponds to the pixel of the second image $I_2$, but \emph{still expressed in the reference frame $\pi_1$ of the first camera}.
This means that the extrinsics $K_2$ and the relative rigid motion $(R^*,\bt^*) \in SE(3)$ from the second camera to the first can be recovered from the analogous equation
$
U_2 \Lambda_2 = K_2 \, (R^*)^{-1} (P_2(\pi_1) - \bt^*).
$
Later, we will discuss an alternative method based on matching point clouds.

The point maps also encode correspondences between image $I_1$ and $I_2$ as it is immediate to tell which 3D points are the same by checking for equality of coordinates given that these are expressed in the same reference frame.
Specifically, to find out which pixels $\bu_i$ in image $I_1$ is the best match for pixel $\bu_j$ in image $I_2$ one simply minimizes the distance between the corresponding 3D points $\| [P_1(\pi_1)]_{:,i} - [P_2(\pi_1)]_{:,j} \|$, which is meaningful as they are both expressed in the same reference frame $\pi_1$.

It is useful to express these correspondences using a matrix notation.
Hence, given two set of $d$-dimensional point descriptors $A^{d\times HW}$ and $B^{d \times HW}$, we define
\begin{equation}\label{e:corresp}
C(A,B) = \operatornamewithlimits{argmin}_{C} \| A - B C \|
\end{equation}
where $C \in \{0,1\}^{HW\times HW}$ is a square binary matrix with exactly one unitary entry along each row, i.e., $C \mathbf{1} = \mathbf{1}$.
Hence, the correspondence matrix from image $I_2$ back to image $I_1$ is simply:
$$
C_{12} = C(P_1(\pi_1), P_2(\pi_2)).
$$


\subsection{Dynamic point maps}%
\label{sec:dynamic-point-maps-theory}

The arguments above break if physical points can move over time, and if the two shots $I_1$ and $I_2$ are not taken simultaneous.
In this case, in fact, a physical point $\p$ will not, in general, found at the same location in the two shots.
Hence, compensating for the camera viewpoint is insufficient in order to establish correspondences.

Recall that our solution is to parametrise points w.r.t.~viewpoint as well as \emph{time}.
Images $I_1$ and $I_2$ come in fact with timestamps $t_1$ and $t_2$, so the coordinate of physical point $p$ are a function of both viewpoint and time, i.e., $\p(t_i,\pi_j)$.

Given the two images, we have four possible combinations of these reference parameters.
Two, i.e., $(t_1,\pi_1)$ $(t_2,\pi_2)$, correspond to the viewpoint and timestamp of image $I_1$ and $I_2$, respectively.
The other two, i.e.,  $(t_1,\pi_2)$, $(t_2,\pi_1)$, correspond to swapping the viewpoint and timestamp between the two images.

Because there are two point clouds $P_1$ and $P_2$, the first corresponding to the pixels in image $I_1$ and the second to the ones in image $I_2$, and each is expressible in any of these references, there are in total eight different outputs.
Since the role of image $I_1$ and $I_2$ is symmetric, it suffices to task the network $\Phi$ to predict four quantities, with the other four obtained by swapping the inputs.
These four predictions are:
\begin{equation}\label{e:core}
(
P_1(t_1,\pi_1),
P_1(t_2,\pi_1),
P_2(t_1,\pi_1),
P_2(t_2,\pi_1)
)
=
\Phi(I_1,I_2).
\end{equation}
Note that all these predictions refer the point clouds to the reference frame $\pi_1$ of the first camera.
We obtain the four complementary predictions for $\pi_2$ by swapping the network's inputs.

\paragraph{Special cases.}

If the images are taken at the same time, then $t_1=t_2$, the predictions
$P_1(t_1,\pi_1) = P_1(t_2,\pi_1)$ and
$P_2(t_1,\pi_1) = P_2(t_2,\pi_1)$ collapse, and model \cref{e:core} reduces to \cref{e:duster} explored by DUSt3R.
Furthermore, if $\pi_1=\pi_2$ as well, then this model reduces to monocular point prediction which, as we have seen above, is related to but more informative than depth prediction.

\subsection{Using \method to solve 3D and 4D tasks}%
\label{sec:reduction}

In this section, we show how the output of the network $\Phi$ can be used to solve a number of basic 3D and 4D problems.

\paragraph{Recovering the camera intrinsics and extrinsics.}

The camera intrinsics $K_1$ can be recovered immediately from equation
$
U_1 \Lambda_1 = K_1 P_1(t_1,\pi_1).
$
The camera extrinsics $K_2$ and the relative rigid motion $(R^*,\bt^*) \in SE(3)$ from the second camera to the first can be recovered from the analogous equation
$
U_2 \Lambda_2 = K_2 \, (R^*)^{-1} (P_2(t_1,\pi_1) - \bt^*).
$
These are the same equations for static point maps, which is possible because we are fixing the time to $t_1$.

\paragraph{Performing motion segmentation.}

To tell whether pixel $\bu_i$ in image $I_1$ corresponds to a physical points that moves with respect to the camera, we can simply check if its coordinates $[P_1(t_1,\pi_1)]_{:,i}$ and $[P_1(t_2,\pi_1)]_{:,i}$ differ or not (see \cref{f:motion_seg}).
We introduce the following compact notation:
given point descriptors $A,B \in \mathbb{R}^{d\times HW}$, we define the mask $M(A,B) \in \{0,1\}^{HW}$ as the vector
$
[M(A,B)]_i = \chi(\|A_{:,i} - B_{:,i}\| > \epsilon),
$
where $\epsilon \geq 0$ is a threshold.
Then, the motion masks in images $I_1$ and $I_2$ are:
\begin{align*}
M_1 &= M(P_1(t_1,\pi_1), P_1(t_2,\pi_1)), \\
M_2 &= M(P_2(t_1,\pi_1), P_2(t_2,\pi_1)).
\end{align*}

\paragraph{Obtaining point correspondences.}

Using \cref{e:corresp}, we can map each pixel in image $I_2$ to the corresponding pixel image $I_1$ via the correspondence matrix:
$
C_{12} = C(P_1(t_1,\pi_1), P_2(t_1,\pi_1)).
$
Note that this also works for dynamic points because the network registers both viewpoint and timestamp, see \cref{f:matching}.

\paragraph{Reconstructing the camera motion.}

The relative rigid motion $(R^*,\bt^*) \in SE(3)$ from the second camera to the first can be recovered from the matching points as
\begin{multline*}
\operatornamewithlimits{argmin}_{R,\bt}
\|
(P_1(t_1,\pi_1) C_{12} -
R P_2(t_1,\pi_2) - \bt)
\operatorname{diag}(\bar M_2)
\|,
\end{multline*}
where the subtraction by $\bt$ is broadcast to all points and $\bar M_2 = \mathbf{1} - M_2$ masks out dynamic pixels, see \Cref{f:cams}.

\paragraph{Reconstructing rigid object motion.}

If $M$ is the mask of a certain object in image $I_1$, then its rigid motion with respect to the reference frame $(t_1,\pi_1)$ can be recovered as (see \Cref{f:obj_track}):
$$
\operatornamewithlimits{argmin}_{R,\bt}
\|
(P_1(t_2,\pi_1) -
R P_1(t_1,\pi_1) - \bt)
\operatorname{diag}(M)
\|.
$$



\section{Experimental details}%
\begin{table}[t]
\centering
\tablestyle{1pt}{1.0}
\resizebox{\linewidth}{!}{
\begin{tabular}{llc|cccc}
\toprule
& Dataset & Motion? & $P_1(t_1, \pi_1)$ & $P_2(t_1, \pi_1)$ & $P_1(t_2, \pi_1)$ & $P_2(t_2, \pi_1)$ \\
\midrule
\multirow{2}{*}{(a)} & Kubric & Yes & \greencheck & \greencheck & \bluecheck & \bluecheck \\
& Waymo  & Yes & \greencheck & \greencheck & \bluecheck & \bluecheck \\
\midrule
\multirow{2}{*}{(b)} & PointOdyssey & Yes & \greencheck & --- & --- & \bluecheck \\
& Spring       & Yes & \greencheck & --- & --- & \bluecheck  \\
\midrule
\multirow{3}{*}{(c)} & ScanNet++  & No & \magcheck & \magcheck & \magcheck & \magcheck \\
& BlendedMVS & No & \magcheck & \magcheck & \magcheck & \magcheck \\
& MegaDepth  & No & \magcheck & \magcheck & \magcheck & \magcheck \\
\bottomrule
\end{tabular}
}
\caption{Training datasets fall into 3 groups. (a) Kubric and Waymo contain dynamic scenes and provide annotations for all 4 point maps. (b) PointOdyssey and Spring only supervise \textit{same-time} point maps. (c) Finally, ScanNet++, BlendedMVS and MegaDepth contain static scenes.}%
\label{tab:datasets}
\end{table}

\label{sec:appendix-experimental-details}

\subsection{Scene and object flow metrics}%
\label{sec:appendix-scene-flow-metrics}
Scene Flow and Object Flow are defined based on the 3D displacement of points in a scene, with and without camera motion. 

\begin{itemize}
    \item \textbf{Scene Flow (SF)} captures the full 3D motion of points, incorporating both object and camera movement:
    \begin{itemize}
        \item \textbf{Forward Scene Flow (SF-F):} 
        $
        P_1(t_2, \pi_2) - P_1(t_1, \pi_1)
        $
        describing how points at \( t_1 \) move to \( t_2 \) under a potentially moving camera.
        \item \textbf{Backward Scene Flow (SF-B):} 
        $
        P_2(t_1, \pi_1) - P_2(t_2, \pi_2)
        $
        mapping how points at \( t_2 \) correspond back to \( t_1 \).
    \end{itemize}

    \item \textbf{Object Flow (OF)} isolates object motion by assuming a fixed camera:
    \begin{itemize}
        \item \textbf{Forward Object Flow (OF-F):} 
        $
        P_1(t_2, \pi_1) - P_1(t_1, \pi_1)
        $
        capturing how points move between frames when viewed from the same camera pose.
        \item \textbf{Backward Object Flow (OF-B):} 
        $    P_2(t_1, \pi_1) - P_2(t_2, \pi_1)
        $
        tracing the movement of points back in time while keeping the viewpoint unchanged.
    \end{itemize}
\end{itemize}
\end{document}